\documentclass[authoryear]{article}
\usepackage{amsmath,amsfonts}
\usepackage{algorithmic}
\usepackage{algorithm}
\usepackage{array}
\usepackage{subcaption}
\usepackage{textcomp}
\usepackage{hyperref}
\usepackage{siunitx}
\usepackage{verbatim}
\usepackage{graphicx}
 \usepackage{natbib}
\usepackage{authblk}

\title{Hybrid Interval Type-2 Mamdani-TSK Fuzzy System for Regression Analysis}
		
		\author[]{Ashish Bhatia}		
		\author[]{Renato Cordeiro de Amorim }		
		\author[]{Vito De Feo }		
		\affil[]{University of Essex, United Kingdom}
        \date{}
\begin{document}
\maketitle
\begin{abstract}
			Regression analysis is employed to examine and quantify the relationships between input variables and a dependent and continuous output variable. It is widely used for predictive modelling in fields such as finance, healthcare, and engineering.  However, traditional methods often struggle with real-world data complexities, including uncertainty and ambiguity. While deep learning approaches excel at capturing complex non-linear relationships, they lack interpretability and risk over-fitting on small datasets. Fuzzy systems provide an alternative framework for handling uncertainty and imprecision, with Mamdani and Takagi-Sugeno-Kang (TSK) systems offering complementary strengths: interpretability versus accuracy. This paper presents a novel fuzzy regression method that combines the interpretability of Mamdani systems with the precision of TSK models. The proposed approach introduces a hybrid rule structure with fuzzy and crisp components and dual dominance types, enhancing both accuracy and explainability. Evaluations on benchmark datasets demonstrate state-of-the-art performance in several cases, with rules maintaining a component similar to traditional Mamdani systems while improving precision through improved rule outputs. This hybrid methodology offers a balanced and versatile tool for predictive modelling, addressing the trade-off between interpretability and accuracy inherent in fuzzy systems. In the 6 datasets tested, the proposed approach gave the best fuzzy methodology score in 4 datasets, out-performed the opaque models in 2 datasets and produced the best overall score in 1 dataset with the improvements in RMSE ranging from 0.4\% to 19\%.
		\end{abstract}
\section{Introduction}
Regression analysis is a fundamental method in statistical modelling and machine learning, employed to examine and quantify the relationships between variables (the inputs) and a continuous dependent variable (the output). This technique serves as a critical tool for predictive tasks, facilitating the analysis of patterns and trends across various fields. Its applications span numerous domains, including finance, healthcare, engineering, and environmental science, where it has driven significant advances in prediction, forecasting, and diagnostic capabilities~\cite{shinde_review_2018}.
Traditional regression techniques, such as linear regression and support vector regression, have proven effective in handling numerous problems. However, these methods often encounter limitations when dealing with the complexities of real-world data. The presence of uncertainty, ambiguity, and linguistic information in datasets poses challenges for traditional approaches that assume precise and well-defined numerical relationships~\citep{maturo_fuzzy_2017}. 
In recent years, the resurgence of neural networks and deep learning has led to considerable progress in various machine learning tasks, including regression. Deep learning models, with their ability to learn complex non-linear relationships from vast amounts of data, have achieved state-of-the-art results in many domains. However, deep learning models lack interpretability, making it difficult to understand the underlying reasoning behind their predictions. In addition to being
un-interpretable, deep learning models, characterized by their large number of parameters, are generally ill-suited for training on small datasets due to the risk of over-fitting and the insufficient representation of data needed to effectively optimize such complex architectures~\citep{rather_breaking_2024}.
Fuzzy logic, introduced by Lotfi A. Zadeh in 1965~\citep{zadeh_birth_1999}, offers a framework for representing and reasoning with uncertain or imprecise information.
By incorporating human-like reasoning and linguistic rules, fuzzy logic systems provide a powerful mechanism for handling problems characterized by vagueness and subjectivity. In the realm of regression, fuzzy approaches have gained prominence due to their ability to effectively model data relationships that are not easily captured by traditional methods~\citep{zadeh_is_2008}.
Among the various fuzzy regression techniques, the Mamdani~\citep{mamdani_experiment_1975} and Takagi-Sugeno-Kang (TSK)~\citep{takagi_fuzzy_1985} approaches are the two main approaches with a long history of application in regression-focused fuzzy systems. Mamdani fuzzy systems, known for their interpretability, explainability and transparency, employ fuzzy IF-THEN rules to map inputs to outputs. Both antecedents (inputs) and consequents (outputs) are represented as fuzzy sets, allowing for the incorporation of linguistic terms and human knowledge into the model. However, the benefits of explainability and interpretability often come at the cost of
model performance in Mamdani systems~\citep{wang_comparison_2014}. One of the main reasons for this is that to improve interpretability, Mamdani systems often use
a limited number of linguistic terms. This coarse granularity can lead to oversimplified models that fail to capture subtle variations in the data, resulting in low regression accuracy~\citep{cordon_historical_2012}.
On the other hand, TSK fuzzy systems, introduced in the 1980s, address the precision limitations sometimes associated with Mamdani systems. It does so by utilizing crisp mathematical functions in the rule consequents, TSK models offer enhanced accuracy in numerical regression tasks. However, this often leads to a lack of explainability and interpretability of the resulting systems~\citep{yen_improving_1998}, defeating the primary purpose of using fuzzy systems, which is generally associated with higher explainability and interpretability.
Fuzzy rule-based systems (FRBSs) have been extensively studied for their ability to handle uncertainty and their application in diverse areas such as classification, control systems, medical predictions, and general modelling tasks. A significant early development in this field was the work of Wang and Mendel~\citep{wang_generating_1991}, proposing a method to generate fuzzy rules directly from numerical data. This data-driven approach laid the groundwork for rule generation techniques, emphasizing interpretability but encountering difficulties in managing complex systems with high accuracy.
Genetic Fuzzy Systems (GFSs) emerged in the early 1990s, integrating Genetic Algorithms (GAs) into FRBSs to automate rule learning and optimization. Hybrid systems like Genetic Fuzzy Neural Networks (GFNNs) were later introduced to enhance accuracy, though often at the cost of the interpretability associated with pure fuzzy approaches.
Advancements in FRBSs continued with the application of evolutionary algorithms to refine the system's Knowledge Base (KB), which comprises the Data Base (DB) and Rule Base (RB). Techniques such as scaling function adjustment and membership function tuning have been used to improve accuracy. For example, GAs have been employed to adjust scaling factors and the shapes of membership functions (e.g., triangular or trapezoidal)~\citep{herrera_genetic_2005,ishibuchi_genetic_2004}. However, fine-tuning membership functions can compromise the meaningful representation of linguistic terms for users. Similarly, RB has been optimized through techniques such as rule selection, pruning, and generation, often employing chromosome designs that balance interpretability and accuracy~\citep{herrera_genetic_2008,herrera_genetic_2005,ishibuchi_genetic_2004}. Evolutionary algorithms, driven by fitness functions, utilize operators like mutation, crossover, and selection to optimize FRBS performance, focusing on objectives such as accuracy and interpretability. These methods have been applied in
areas such as control systems, classification, and forecasting~(\citep{cordon_historical_2012}.
Efforts to optimize the DB have also contributed to improving fuzzy systems. For instance, Magdalena introduced non-linear scaling functions to enhance membership function's sensitivity in key input regions~\citep{magdalena_fuzzy_1997}. Further, methods such as parent-centric recombination~\citep{deb_real-coded_2002} and simulated annealing approaches~\citep{cordon_analysis_2000} explored optimization avenues. The Best-Worst Ant System (BWAS)~\citep{cordon_analysis_2000} combined ant colony optimization (ACO) principles with evolutionary algorithms to enhance system performance~\citep{magdalena_fuzzy_1997}.
Other methodologies, such as the Cooperative Rule (COR) method~\citep{casillas_cor_2002}, have aimed to reduce rule conflicts
in FRBSs. Unlike traditional methods that optimize rules individually, COR adopts a holistic approach, enhancing rule interaction to improve accuracy and reduce conflicts.
Recent advances have focused on hybrid models that combine RB and DB optimization to balance accuracy and interpretability. Approaches like Granularity + Membership Function (Gr-MF) and GA-WM (Wang-Mendel Method + Global tuning of membership functions using a genetic algorithm) use genetic algorithms to optimize fuzzy rule learning and the tuning of membership functions, leading to systems with improved accuracy and adaptability. For instance, the GA-COR method integrates genetic algorithms with the COR framework to optimize rule sets and database parameters, creating more compact and interpretable models. The WM and genetic learning (WM + GL) approach incorporates linguistic 2-tuples for lateral tuning of membership functions, enabling more efficient search space exploration~\citep{alcala_genetic_2007}.
Despite these advances, challenges persist in achieving a balance between accuracy and interpretability in GFSs and FRBSs. Many evolutionary algorithms prioritize accuracy, leading to larger and more complex rule sets that are less interpretable. Fine-tuning membership functions and rule parameters to improve accuracy may result in models that are overfitted to the training data, reducing their generalizability. Conversely, methods emphasizing interpretability, such as traditional Mamdani models and COR methodologies, tend to generate simpler, more transparent rule sets. While these models are easier to understand, they
often lack the precision needed for complex applications requiring detailed adjustments to membership functions and rules. These limitations highlight the need for hybrid approaches capable of maintaining an effective balance between interpretability and accuracy, a key focus of current research in this domain.
Despite the numerous advantages of fuzzy approaches in regression, it is still challenging to balance the interpretability and accuracy of the developed fuzzy regression systems. This paper contributes to the field of fuzzy regression modelling by developing a novel system, Hybrid IT2 Mamdani TSK fuzzy system (HIT2-MTSK), that addresses these challenges as follows:
\begin{itemize}
\item Integrates the linguistic interpretability of Mamdani systems with the numerical precision of TSK models.
\item Introduces a new rule type which has two components - a fuzzy component and a function component providing a crisp output value for the rule. Instead of pointing to the centroid of the consequent fuzzy set, the output in the proposed system is derived from an equation involving the antecedent features.
\item This paper presents two dominance types for the rules -- one associated with the fuzzy consequent and another with the crisp consequent.
\end{itemize}

Hence, this work aims to develop a robust framework capable of accurately handling complex, real-world datasets while preserving the interpretability and adaptability of the developed fuzzy regression systems. This paper aims not only to advance the state-of-the-art in fuzzy regression but also to contribute to the broader field of machine learning by providing a more versatile and robust tool for predictive modelling. The proposed technique was evaluated on a range of benchmark datasets to demonstrate its effectiveness and potential for practical applications. In the 6 datasets tested, the proposed approach gave the
best fuzzy methodology score in 4 and best overall score in 1 dataset. The improvements in Root Mean Squared Error (RMSE) ranged from 0.4\% to 19\%.
In Section II, we present an overview of Type-II Fuzzy Rule Based Systems (FRBS). Section III presents a brief overview of ACO. In Section IV, our proposed approach for the HIT2-MTSK is presented. In section V, the experiments conducted to test the efficacy of the HIT2-MTSK approach are presented followed by conclusion and future research direction in section VI.
\section{Overview on Type-2 Fuzzy Rule Based Systems}
\begin{figure}
  \centering
  \includegraphics[width=2.5in]{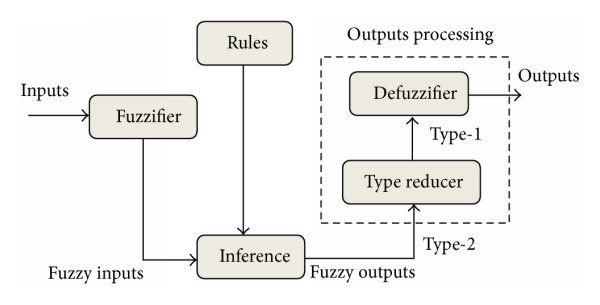}
  \caption{Type-2 Fuzzy Logic System} 
  \label{IT2FLS} 
\end{figure}

Fuzzy rule-based systems (FRBSs) are a foundational approach in fuzzy set theory, allowing for the representation and reasoning with uncertain knowledge. These systems extend classical rule-based systems by incorporating fuzzy sets and fuzzy logic, enabling them to handle the inherent vagueness and imprecision of real-world problems.
FRBSs are characterized by two primary components: a knowledge base and a processing structure. The knowledge base stores the system's knowledge, including fuzzy partitions, which define linguistic terms and their associated membership functions; a rule base, comprising fuzzy IF-THEN rules that capture relationships
between variables; and scaling functions, which map between the system and fuzzy domains~\citep{magdalena_fuzzy_2015}.
The processing structure carries out the reasoning process, transforming crisp inputs into fuzzy sets through a fuzzification interface, applying the fuzzy rules in an inference engine, and converting the resulting fuzzy outputs back into a crisp value via a defuzzification interface. In some systems, input and output scaling functions ensure proper domain mapping.
FRBSs offer a robust framework for handling uncertainty and have been successfully applied in various domains, including control systems, pattern recognition, and decision support systems. Their ability to represent and reason with linguistic information makes them particularly suitable for problems where human knowledge and experience play a crucial role.
Traditional Mamdani-type FRBSs employ type-1 fuzzy sets, where the membership degree of an element in a fuzzy set is a crisp value between 0 and 1. However, in situations with high uncertainty or ambiguity, type-2 fuzzy sets can provide a more nuanced representation~\citep{karnik_type-2_1999}. Type-2 fuzzy sets have fuzzy membership grades themselves, allowing for a degree of uncertainty in the membership function. This is represented by a footprint of uncertainty (FOU), which encompasses the range of possible membership functions. Type-2 Mamdani-type FRBSs can handle uncertainties more effectively, leading to improved performance in certain applications~\citep{mendel_uncertain_2017}.
Interval Type-2 fuzzy sets simplify a Type-2 fuzzy set by representing the membership function as a bounded interval at each point in the universe of discourse. This interval accounts for uncertainty in the definition of the membership function, providing additional flexibility compared to Type-1 systems. The extra degree of freedom enables IT2-FRBSs to model systems where data or expert knowledge is imprecise, conflicting, or noisy~\citep{mendel_uncertain_2017, sanz_compact_2015}.
An IT2-FRBS retains the general structure of a traditional FRBS but incorporates Interval Type-2 fuzzy sets within its knowledge base and reasoning process. The structure of a rule in an IT2-FRBS is
\begin{multline}
IF\  x_1\  is\ F_1\   AND \ldots AND\ x_n\   is\ F_n\ THEN\ y\ is\ G,
  \label{mamdanirule}
\end{multline}
where \(x_{1},\ldots, x_{n}\) are the \(n\) input variables, \(F_{1}, \ldots, F_{n}\) are the IT2 fuzzy sets associated with each input variable, \(y\) is the output variable, and \(G\) is an IT2 fuzzy set associated with the output \(y\). Fig. \ref{IT2FLS} shows the structure of a type-2 fuzzy logic system.
TSK fuzzy systems offer a distinct approach to fuzzy regression with a focus on mathematical efficiency and functional representation. They have gained popularity in applications such as control, system identification, and data analysis~\citep{liang_introduction_1999}. TSK fuzzy systems also employ ``IF-THEN'' rules, but the consequents are linear or non-linear functions of the input variables. A typical TSK rule might take the form
\begin{multline}
IF\ input1\ is\ LOW\ AND\ input2\ is\ HIGH\ \\
THEN\ output\ = a*input1\ + b*input2 + c.
  \label{eq:tskrule}
\end{multline}
Here, ``input1'',  ``input2'' and ``output'' are variables, while ``low'' and ``high'' are linguistic terms associated with fuzzy sets defined over the
respective universes of discourse. The coefficient terms - \(a\), \(b\), and \(c\) are constants.
The inference process in a TSK fuzzy system involves~\citep{liang_introduction_1999}:

\begin{enumerate}
\def\labelenumi{\arabic{enumi}.}
\item
  Fuzzification\textbf{:} Similar to Mamdani-type FRBSs, input values are fuzzified to determine their membership degrees in the antecedent fuzzy sets.
\item
  Rule Application: The firing strength of each rule is calculated based on the fuzzified inputs, typically using a t-norm operator.
\item
  Output Calculation: The output of each rule is calculated using the consequent function and the corresponding firing strength.
\item
  Weighted Averaging: The overall output is obtained by taking the weighted average of the outputs from each rule, where the weights are the normalized firing strengths.
\end{enumerate}
\begin{equation}
y_{final} = \frac{\sum_{i = 1}^{M}{\omega_{i}y_{i}}}{\sum_{i = 1}^{M}\omega_{i}}.
\label{eq:weighted_defuzz}
\end{equation}
Liang and Mendel have discussed three types of TSK system architectures based on antecedent and consequent configurations while also incorporating interval type-2 sets in the TSK models~\citep{liang_introduction_1999}. These are,

\begin{itemize}
\item
  Type-2 antecedents, Type-1 fuzzy consequent parameters.
\item
  Type-2 antecedents, crisp consequent parameters.
\item
  Type-1 antecedents, Type-1 fuzzy consequent parameters.
\end{itemize}
\section{A brief overview of Ant Colony Optimization}
ACO is a meta-heuristic algorithm inspired by the foraging behaviour of ant colonies~\citep{dorigo_ant_1996}. It is typically used to find approximate solutions to combinatorial optimization problems. The core ideas of the algorithm are:
\begin{itemize}
    \item Pheromone Trails: Real ants deposit a chemical substance called pheromone as they move. The concentration of pheromone on a path indicates its attractiveness. Other ants are more likely to follow paths with higher pheromone levels.
    \item Stigmergy: Ants indirectly communicate by modifying their environment with pheromone. This indirect communication mechanism allows the colony to collectively find efficient paths.
    \item Probabilistic Decisions: In ACO algorithms, artificial ants probabilistically choose their next move based on the pheromone concentration and heuristic information (e.g., distance).
    \item Evaporation Pheromone: Trails evaporate over time, preventing premature convergence to suboptimal solutions and encouraging exploration of new paths.
    \item Reinforcement: Shorter, more optimal paths will be traversed more frequently, leading to a higher pheromone concentration and further reinforcing these paths.
\end{itemize}
The generic structure of the ACO algorithm is as given below.

\begin{itemize}
\item Parameters:
      \begin{itemize}
        \item $num\_ants$: Number of artificial ants
        \item $num\_iterations$: Maximum number of iterations
        \item $\alpha$: Influence of pheromone ($\alpha \ge 1$)
        \item $\beta$: Influence of heuristic information ($\beta \ge 0$)
        \item $\rho$: Pheromone evaporation rate ($0 < \rho \le 1$)
        \item $Q$: Pheromone deposit constant
        \item $initial\_pheromone$: Initial pheromone level on all edges (small positive value)
      \end{itemize}
    \item Initialization: The pheromone trails on all possible connections (edges in the graph) are initialized to a small positive value. This ensures that all paths have a non-zero initial probability of being chosen. The \texttt{best\_solution} and its cost are initialized to keep track of the best solution found so far.
    \item Iteration: The main loop runs for a predefined number of \texttt{num\_iterations}.
    \item Ant Solution Construction: In each iteration, each of the \texttt{num\_ants} constructs a solution. Each ant starts from a designated starting node.
    In each step, the ant probabilistically chooses the next node to visit from its unvisited neighbours.
    \begin{itemize}
        \item Probability Calculation: The probability of choosing a neighbour depends on two factors:
        \begin{itemize}
            \item Pheromone Level: Edges with higher pheromone concentration are more attractive. The influence of pheromone is controlled by the parameter $\alpha$.
            \item Heuristic Information: This is problem-specific information that guides the search (e.g., for TSP, it is typically the inverse of the distance between two cities, making shorter distances more attractive). The influence of the heuristic is controlled by the parameter $\beta$.
        \end{itemize}
        The probabilities for all unvisited neighbours are calculated and then normalized to form a probability distribution.
    \end{itemize}
    The ant continues this process until a complete solution is constructed. The cost of the constructed solution is calculated. If the current ant's solution is better than the \texttt{best\_solution} found so far, the \texttt{best\_solution} and \texttt{best\_solution\_cost} are updated.
    \item Pheromone Update:
    \begin{itemize}
        \item Evaporation: The pheromone on all edges is reduced by a factor of $\rho$. This simulates the natural evaporation of pheromone and prevents premature convergence to a suboptimal solution by making previously visited paths less attractive over time.
        \item Pheromone Deposit: Ants that have found good solutions deposit pheromone on the edges they traversed. The amount of pheromone deposited is usually proportional to the quality of the solution (e.g., inversely proportional to the solution cost). The parameter $Q$ is a constant that influences the amount of pheromone deposited.
    \end{itemize}
    \item Termination: After the specified number of iterations, the algorithm terminates, and the \texttt{best\_solution} found is returned.
\end{itemize}
The specific implementation used in this research is discussed in the next section. 

\section{The Proposed Method}
\begin{figure}
  \centering
  \includegraphics[width=2.5in]{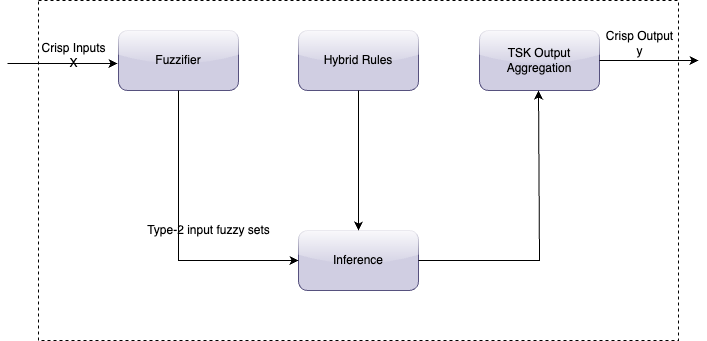}
  \caption{Structure of Hybrid Mamdani-TSK System} 
    \label{hybrid_Struct} 
\end{figure}

The HIT2-MTSK system integrates a interval type-2 fuzzy system with a hybrid Mamdani-TSK framework to address regression challenges by combining interpretability and precision. The steps involved in the HIT2-MTSK model are (Fig.\ref{hybrid_Struct}):

\begin{itemize}
\item Fuzzification of Input and Output with Interval Type-2 Fuzzy Sets.
\item Rule Generation with Mamdani and TSK Components.
\item Rule Base selection using ACO.
\item Defuzzification.
\end{itemize}

\begin{figure}
\centering
 \includegraphics[width=2.5in]{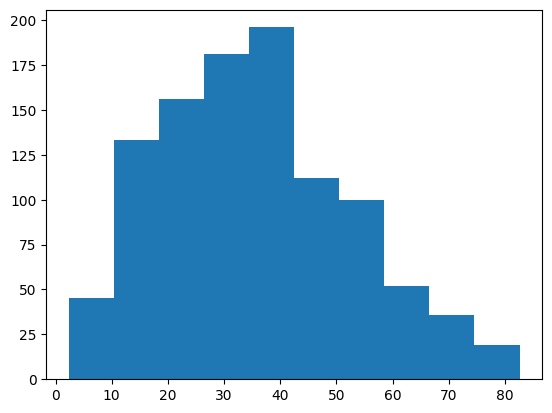}
 \caption{Distribution for Concrete Compressive Strength}
 \label{comp_str}
\end{figure}

\begin{figure}
\centering
 \includegraphics[width=2.5in]{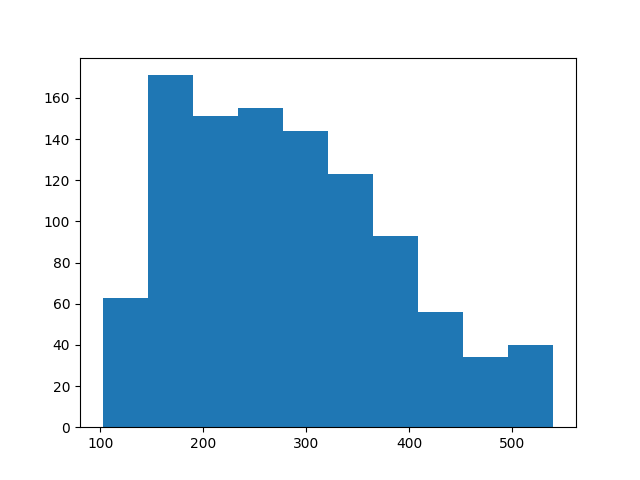}
 \caption{Distribution for Concrete Cement}
 \label{cement}
\end{figure}

\begin{figure}
\centering
 \includegraphics[width=2.5in]{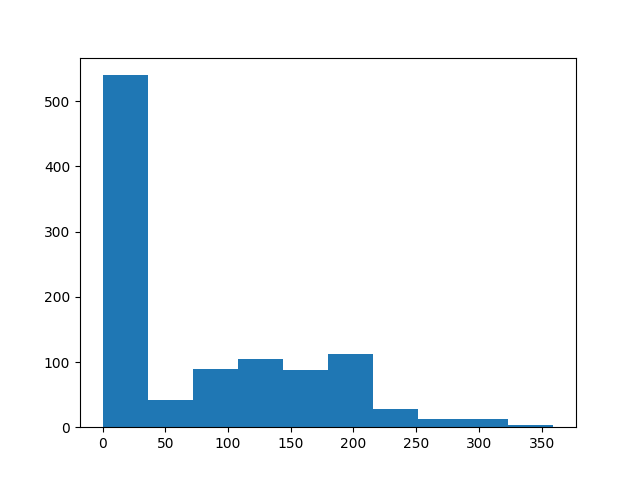}
 \caption{Distribution for Blast Furnace Slag}
 \label{bfslag}
\end{figure}

In this section, we use a running example with two sample features (cement, blast furnace slag) from the Concrete dataset to explain the relevant concepts. The distribution of the regression target (compressive strength) and the selected features for this dataset are as shown in Fig. \ref{comp_str}, \ref{cement} and \ref{bfslag}. Three fuzzy sets -- Low, Medium and High, were used to fuzzify the variables as shown in Fig.\ref{sch_it2}.

\subsection{Fuzzification with Interval Type-2 Fuzzy Sets}

Inputs and outputs are fuzzified using interval type-2 fuzzy sets, which better represent uncertainty by allowing membership values to vary within an interval. This is defined as
\begin{equation}
\tilde{F} = \int_{x \in X} \left[ \int_{u \in [\underline{\mu}(x), \overline{\mu}(x)]} 1/u \right] /x,
\end{equation}
where \(\underline{\mu}(x)\) and \(\ \bar{\mu}(x)\ \)are the lower and upper membership degrees. Interval type-2 sets handle
uncertainty more flexibly than type-1 fuzzy sets, enhancing robustness for real-world applications.

To enhance interpretability and explainability, the input and output variables are usually divided into three fuzzy sets, corresponding to the linguistic terms Low, Medium, and High. This ensures that the partitions align with human understanding and provide meaningful linguistic descriptions of the data. However, more partitions can be used, if appropriate.

\begin{figure}
\centering
 \includegraphics[width=2.5in]{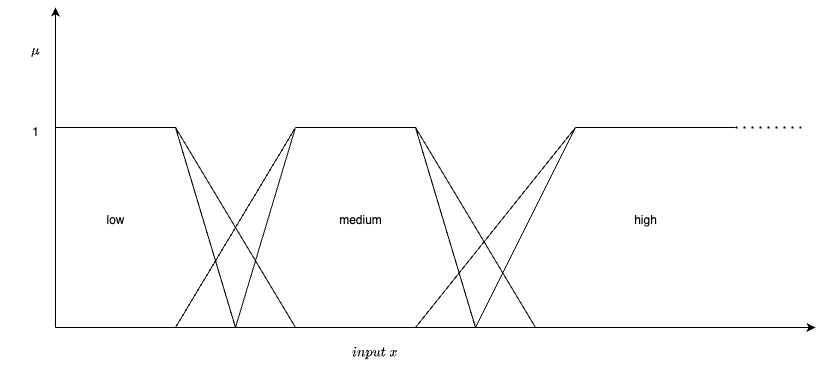}
 \caption{Schema for Interval Type-II Fuzzy Sets}
 \label{sch_it2}
\end{figure}

The fuzzy sets are generated based on the data distribution. The extreme left and extreme right fuzzy sets are modelled as interval type-2 left shoulder and interval type-2 right shoulder respectively. All other fuzzy sets are modelled as interval type-2 trapezoidal fuzzy sets as shown in Fig.\ref{sch_it2}. To ensure that the linguistic terms Low, Medium, and High reflect meaningful data characteristics, the parameters for each fuzzy set are determined based on the underlying variables' distribution.

\begin{figure}
\centering
\includegraphics[width=2.5in]{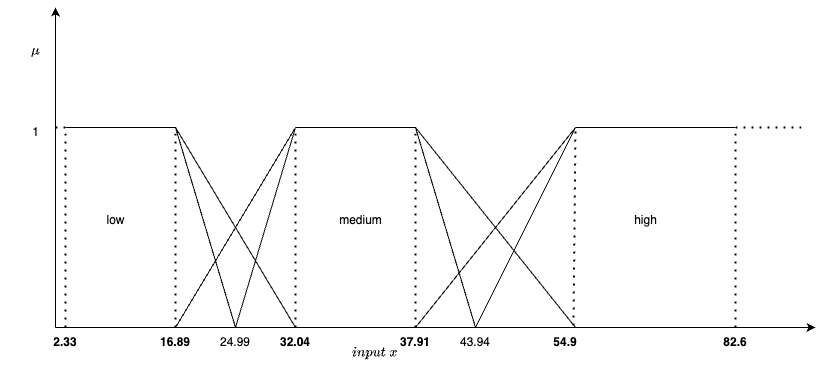}
\caption{Significant Fuzzy Set Points for Concrete Strength}
\label{concrete_points}
\end{figure}

For the Concrete Compressive Strength in our example, the lower and upper bound for each fuzzy set (where the support is greater than zero) are, Low: {[}2.33, 32.04{]}, Medium: {[}16.89, 54.9{]} and High: {[}37.91, 82.6{]} as per the points highlighted in Fig. \ref{concrete_points}.

\subsection{Rule Generation with Hybrid Components and Rule Dominance}
\label{rule-generation-with-hybrid-components-and-rule-dominance}

Each rule in our HIT2-MTSK methodology has two components for the consequent:

\begin{enumerate}
\def\labelenumi{\arabic{enumi}.}
\item
  Fuzzy Component: This is the same as the Consequent in a Mamdani FRBS -- i.e. each rule will point to an output fuzzy set, as shown in Eq. (\ref{mamdanirule}).
\item
  TSK Function Component: A crisp output is provided using a \(n\)-order polynomial equation. However, there are two notable departures from the traditional TSK methodology.
  \begin{itemize}
  \item
    The equation is only trained on subset of training samples (rows) which have positive firing-strength for the fuzzy component, and
  \item
    the output from this equation is forced to be bound within the values of the associated fuzzy component.
  \end{itemize}
\end{enumerate}

The function component can range from a simple linear function $(n=1)$ to polynomials of higher degree $(n>1)$ for non-linear mapping. For instance, for inputs \(x_{1}\) and \(x_{2}\)\hspace{0pt}, and \(n\) set to 2, the function component takes the form:
\begin{equation}
y = w_{0} + w_{1}x_{1} + w_{2}x_{2} + w_{3}x_{1}^{2} + w_{4}x_{2}^{2} + w_{5}x_{1}x_{2},
\end{equation}
where \(w_{0},\ldots,w_{5}\) \hspace{0pt} are coefficients learned from the data. The TSK function component allows the model to capture complex non-linear relationships and makes the prediction system much more accurate when compared to other fuzzy approaches as discussed in Section \ref{experiments-and-results}.

\begin{figure}
\centering
\includegraphics[width=2.5in]{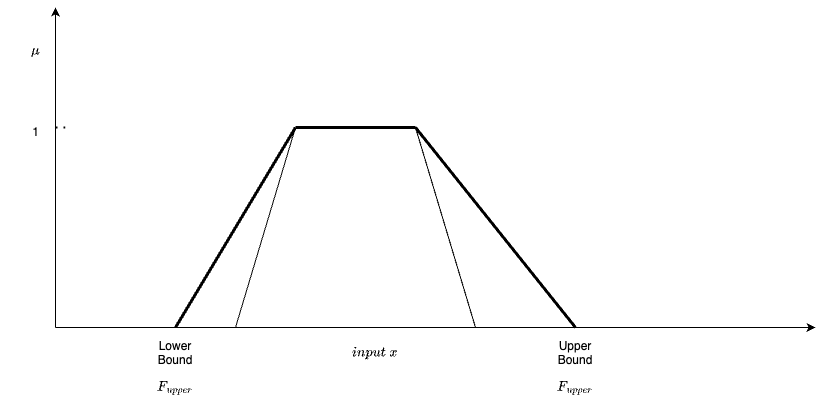}
\caption{Lower and Upper Bounds of a Trapezoid Type-2 Fuzzy set}
\label{bounds}
\end{figure}

To ensure that the meaning of the Mamdani consequent is maintained, we need to make sure that the output from the TSK component is always within the bounds of the consequent fuzzy set. Therefore, the crisp output from a rule is constrained within the fuzzy bounds. That is,
\begin{equation}
y_{\text{output}} \in \left\lbrack \text{LowerBound}\left( F_{upper} \right),\text{UpperBound}\left( F_{upper} \right) \right\rbrack,
\end{equation}
where \(F_{upper}\) is the upper type-1 fuzzy set for the underlying interval type-2 fuzzy set as shown in Fig.\ref{bounds}. This maintains interpretability for the underlying rule by keeping the output within the bounds of the underlying fuzzy set and prevents unreasonable extrapolations.

In our concrete dataset example, for variables \emph{concrete\_cement} and \emph{concrete\_blast\_furnace\_slag,} a sample rule generated in our models is
\begin{equation}
\begin{aligned}
& \text{IF cement is High}  \text{ AND blast\_furnace\_slag is High} \\
& \text{THEN } \big( \text{compressive\_strength\_High}, \\
& \quad y = 0.3 \cdot \text{cement} - 0.6 \cdot \text{blast\_furnace\_slag} \\
& \quad - 5.29 \times 10^{-4} \cdot (\text{cement})^2 \\
& \quad + 1.68 \times 10^{-3} \cdot \text{cement} \cdot \text{blast\_furnace\_slag} \\
& \quad + 2.82 \times 10^{-4} \cdot (\text{blast\_furnace\_slag})^2 \big),
\end{aligned}
\label{cement_eq}
\end{equation}
where the coefficients have been generated from the data. The consequent of rule shown in Eq. (\ref{cement_eq}) is a 2-tuple. The first part of the tuple is the fuzzy set that the rule points to (\emph{High} in the example shown), the second part of the tuple is the TSK function equation which is used to derive the output within the bounds of the fuzzy set to which the rule points. So, in this example, the output from the equation will always be bound within the lower and upper bounds of the Fuzzy set `High' of the Concrete Compressive Strength.

As an example, if Cement value is \num{375} and Blast Furnace Slag is \num{300} -- both values are in the `High' fuzzy set of their features which means that our sample rule will be triggered -- the output from our sample rule, based on the equation shown in Eq. (\ref{cement_eq}) is 72.62, which is inside the lower and upper bound of the High fuzzy set for Compressive Strength and hence does not need changing. Let us assume there was a case in which the rule output was 85 -- i.e. outside the upper bound of 82.6, in this case the rule output would be changed to 82.6 and similarly if there was a case in which the rule output was lower than 37.91, then this would be changed to 37.91.

For this research, 2 variants of the rule types have been used. The first variant sets \(n\) to 2 (i.e., a second order TSK equation and the second variant sets \(n\) to 3 (a third order TSK function component). Since the rules have a traditional Mamdani consequent too, the interpretability of the rule and the results do not depend on the order of the TSK component, which means that we can increase the complexity of the TSK component if desired without any adverse impact on the explainability and interpretability of the rule.

\subsection{Rule Weights}
In a novel departure from other regression techniques, our approach calculates two weights for each rule -- both derived from the training data. The first technique is based on~\citep{sanz_compact_2015} and calculates the rule weight based on the fuzzy support and confidence of the rule. The second weight is based on the accuracy of the TSK rule calculated on the training data.
\begin{itemize}
\item
  Fuzzy Rule Weight: This method calculates fuzzy dominance for rules by combining support and confidence values derived from the fuzzy membership intervals of antecedents and consequents. The steps are as follows.
  \begin{itemize}
  \item
    Calculate Firing Strength (FS): The function starts by determining the firing strength of the rule's antecedent for each data instance. Each premise in the antecedent contributes to the firing strength using a chosen T-norm (e.g., minimum or product). For a given data instance $x$ and a rule with an antecedent having $n$ premises, where each premise $A_i$ is a Type-2 fuzzy set with membership grade $[\mu_{A_{i1}}(x), \mu_{A_{i2}}(x)]$, the firing strength interval $[f_1(x), f_2(x)]$ is calculated using a t-norm $T$,
    \begin{equation}
\begin{aligned}
[f_1(x), f_2(x)] &= T([\mu_{A_{11}}(x), \mu_{A_{12}}(x)], ..., \\
&\qquad [\mu_{A_{n1}}(x), \mu_{A_{n2}}(x)]).
\end{aligned}
 \end{equation}
  \item
    Support of Antecedent and Rule: The support of antecedent is the fuzzy weighted  instances in the input space where the fuzzy membership function of the antecedent has non-zero membership values. For support of the Rule: The firing strength is combined with the consequent's membership interval values to compute the rule's support. The support and confidence Equations (\ref{supp_eq}) and (\ref{conf_eq}) relate to the $j^{th}$ rule with Antecedent $A_j$ and consequent fuzzy set $\tilde{C}_j$ for a dataset with $N$ instances with $x_p =(x_{p1},\ldots,x_{pn})$ and $p = 1, \ldots ,N$.
    \begin{equation}
    \text{Support}(A_j \rightarrow \tilde{C}_j) = \frac{1}{|N|} \sum_{p=1}^{N} \mu_{A_j}(x_p) \cdot \mu_{C_j}(y_p).
    \label{supp_eq}
	\end{equation}

  \item
    Fuzzy Confidence: The rule's confidence is the ratio of its support to the antecedent's support.
    \begin{equation}
    \text{Confidence}(A_j \rightarrow \tilde{C}_j) = \frac{\sum_{p=1}^{N} \mu_{A_j}(x_p) \cdot \mu_{C_j}(y_p)}{\sum_{p=1}^{N} \mu_{A_j}(x_p)}.
    \label{conf_eq}
    \end{equation}
  \item
    Dominance Calculation: Dominance or the weight of the rule is calculated taking into account both the fuzzy support and fuzzy
    confidence of the rule. For example, the dominance can be calculated as
    \begin{equation}
        D = [S_{Rule\_lower} \cdot C_{lower}, S_{Rule\_upper} \cdot C_{upper}].
     \end{equation}
  \end{itemize}
\end{itemize}
\begin{quote}
This ensures the fuzzy weight reflects both how well the rule fits the data (confidence) and its prevalence (support). The main role of fuzzy rule weights is at the time of generation of a viable universe of rules.
\end{quote}
\begin{itemize}
\item
  Error based Dominance: At the time of rule generation, each rule is evaluated using its performance on training data, quantified as its RMSE over applicable instances, wherever the firing strength of the rule is non-zero. The error-based dominance of a rule is taken as
\end{itemize}
\begin{equation}
\text{Dominance}_{r^{i}} = \frac{1}{1 + r^{i}},
\label{dom_error}
\end{equation}
\begin{quote}
where \(r^{i}\) is the RMSE of the i\textsuperscript{th} rule. Rules with lower RMSE have higher dominance values, allowing them to
contribute more significantly during the rule selection process and at time of inference. This ensures the dynamic prioritization of accurate rules.
\end{quote}
In our example, the rule discussed in Eq. (\ref{cement_eq}) has two weights, the fuzzy weight is {[}0.436, 0.457{]} and the TSK component dominance is 0.065 -- which is calculated using its RMSE with Eq. (\ref{dom_error}). The RMSE for this rule is
14.3 and the error-based dominance can be calculated from this as
\begin{equation}
\text{Dominance}_{r^{sample}} = \frac{1}{1 + 14.3\ } = 0.065.
\end{equation}

\subsection{Subset Selection Using ACO}
\begin{figure}
\centering
\includegraphics[width=2.0in]{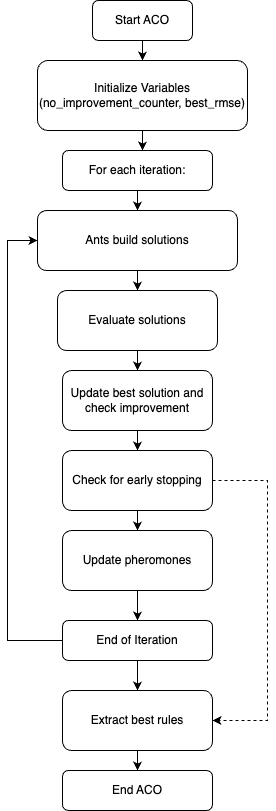}
\caption{Ant Colony Optimisation Overview}
\label{aco}
\end{figure}

After rule generation, a subset of rules is selected using Ant Colony Optimization (ACO). ACO optimizes the trade-off between model compactness and accuracy by iteratively refining a subset of rules. The fitness function evaluates rule subsets based on the combined RMSE of selected rules on training data with an optimization goal of minimizing the RMSE on the training/validation dataset. The ACO algorithm is inspired by the foraging behaviour of ants, who use pheromones to mark paths leading to food sources.
The optimization procedure starts with an initial set of rules, and each ant in the colony constructs a solution by selecting a subset of these rules. The ants evaluate the performance (RMSE) of their solutions on a training dataset. Based on the performance, they deposit pheromones on the rules they selected. Rules that lead to better solutions receive more pheromones, making them more likely to be selected in future iterations. The pheromone levels on the rules evaporate over time, and the process repeats for a fixed number of iterations. The algorithm keeps track of the best solution (set of rules) found so far and its corresponding RMSE. The full process is shown in Fig. \ref{aco}. ACO is computationally efficient, making it well-suited for high-dimensional datasets, while ensuring the selected rule subset retains high predictive performance. The ACO implementation in this research follows the following steps.
\begin{itemize}
\item The algorithm runs multiple iterations until a maximum number of iterations is reached or the score stops improving for a pre-defined number of iterations. 
\item Each ant constructs a solution (a subset of rules) probabilistically using  selection probabilities for each rule by balancing two factors: (1) historical success (represented by ``pheromone'' values) and (2) individual rule accuracy (prioritising rules with lower errors). The system then picks a group of rules within a predefined size range, favouring those with higher probabilities. 
\item The solution is evaluated using RMSE on training and validation data.
\item Identifies the best solution (lowest RMSE) in the current iteration.
\item Updates the global best solution if an improvement is found.
\item If no improvement occurs for a set number of iterations (patience), the loop breaks early.
\item Adjusts pheromone levels based on solution quality.
\item Log the current iteration and best RMSE.
\item After completion, the best-performing rules are extracted from the best solution.
\end{itemize}

\subsection{Defuzzification Using TSK Weighted Mean}

The final output for a given instance is calculated using the weighted mean approach from the TSK model as per Eq. (~\ref{eq:weighted_defuzz}).

\section{Experiments and Results}\label{experiments-and-results}
To evaluate the performance and robustness of our HIT2-MTSK regression approach, we conducted experiments on six benchmark datasets obtained from the KEEL repository~\citep{alcala-fdez_keel_2011}: Concrete Compressive Strength, Diabetes, ELE-2, Mortgage, Treasury and Wankara. These datasets span diverse domains, offering a comprehensive testing and benchmarking comparison for the HIT2-MTSK approach. Below, we provide an overview of these datasets, focusing on their attributes, target variables, and the regression task.
\begin{enumerate}
\def\labelenumi{\arabic{enumi}.}
\item
  Concrete Compressive Strength Dataset: This dataset consists of \num{1030} instances and 8 input variables, including components like cement, water, and aggregates, along with a single target variable, compressive strength. Predicting compressive strength is crucial in civil engineering applications due to its importance in structural integrity. The dataset presents non-linear relationships between inputs and the target.
\item
  Diabetes Dataset: The dataset includes \num{768} records with 2 input attributes. The target variable represents a quantitative measure of diabetes progression. This medical dataset is widely used in predictive modelling for healthcare and presents the challenge of understanding physiological interactions in a regression framework.
\item
ELE-2 Dataset: With \num{11105} instances and 4 predictor variables, this dataset addresses energy load forecasting, a critical task for energy resource management.
\item
  Mortgage Dataset: Comprising \num{1045} instances and 14 variables, this dataset involves predicting the risk of loan delinquency based on borrower characteristics and loan details. It represents a financial domain use case, where accurate predictions are essential for credit risk assessments.
\item
  Treasury Dataset: This dataset contains \num{956} records and 16 financial attributes derived from time-series data. The target variable is a yield rate metric, relevant for economic modelling and investment decision-making.
\item
  Wankara Dataset: This dataset contains \num{1609} records and contains the weather information of Ankara from 01/01/1994 to 28/05/1998. The output to be predicted is the daily mean temperature based on 4 input variables.
\end{enumerate}

\begin{quote}
The KEEL website provides 5-fold cross-validation data for each dataset and the same has been used to compare the average accuracy with the benchmark results provided. The benchmark algorithms as used by the original authors are~\citep{alcala-fdez_keel_2011, alcala_genetic_2007}:
\end{quote}
\begin{itemize}
\item
  MP - Multilayer perceptron.
\item
  SMOreg - Sequential Minimal Optimization.
\item
  WM - Wang and Mendel's algorithm.
\item
  CHV - Cordon, Herrera and Villar's algorithm.
\item
  GLD-WM- Gr. + Global lateral parameters + RB by WM.
\end{itemize}
\subsection{Model Variants}\label{model-variants}

We evaluate two configurations of the HIT2-MTSK model. These are

\begin{figure}
\centering
\includegraphics[width=2.5in]{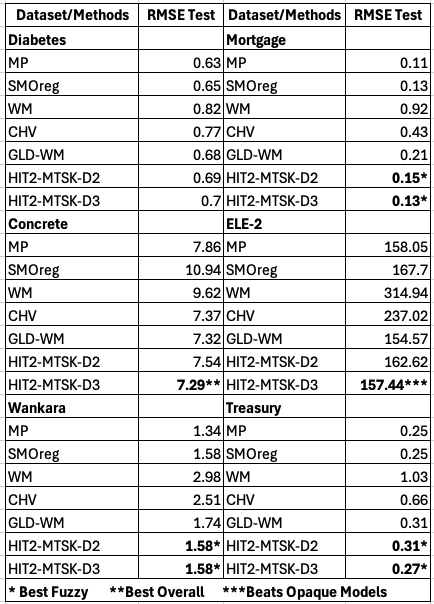}
\caption{Results Comparison against Benchmark Datasets}
\label{results}
\end{figure}

\begin{itemize}
\item
  \textbf{HIT2-MTSK-D2}: TSK function of degree 2.
\item
  \textbf{HIT2-MTSK-D3}: TSK function of degree 3.
\end{itemize}
The results, discussed next, highlight the performance of our hybrid fuzzy approach in achieving accurate predictions while balancing
interpretability. 
Fig.\ref{results} shows the results from the two approaches used and the comparative performance against the benchmark
algorithms. Out of the 6 datasets, the HIT2-MTSK approach is the best fuzzy or overall method in 4 datasets. That is, 66.66\% of datasets, which is a significant achievement. Our approach outperforms the opaque models in 2 datasets and is the best overall
result in 1 dataset, which is also important given that the models from our approach are fully explainable and that the results are achieved without any compromise on the explainability of the rules or the collective rule base even when making use of more accurate rules. 
The performance of the two HIT2-MTSK models varies across datasets, demonstrating their strengths in different domains. On the Concrete Dataset, HIT2-MTSK-D3 is the top performer overall, achieving the lowest RMSE of 7.29, outperforming both fuzzy and non-fuzzy benchmarks like GLD-WM (7.32) and MP (7.86). However, on the Diabetes Dataset, HIT2-MTSK-D3 (RMSE: 0.80) and HIT2-MTSK-D2 (0.79) lag slightly behind non-fuzzy methods such as MP (0.63) and SMOreg (0.65), highlighting a need for methodological refinements to address this gap. Interestingly, the HIT2-MTSK-D2 variant performs slightly better than HIT2-MTSK-D3, suggesting that
the more complex rules are overfitting and are not needed. For the ELE-2 Dataset, HIT2-MTSK-D3 achieves a competitive RMSE of 189.28, significantly better than fuzzy baselines like WM and CHV but still trailing non-fuzzy models like MP (158.05), indicating room for improvement. 

In financial datasets, HIT2-MTSK-D3 performs well. On the Mortgage Dataset, it delivers the best fuzzy model performance with an RMSE of 0.13, surpassing HIT2-MTSK-D2 (0.15) and outperforming all other fuzzy methods (e.g., WM: 0.92) The gap between the best non-fuzzy approach, MP(0.11) is also not very large. Similarly, on the Treasury Dataset, HIT2-MTSK-D3 achieves the lowest fuzzy model RMSE (0.27) and rivals non-fuzzy approaches like MP (0.25). In the Wankara Dataset HIT2-MTSK-D2 and HIT2-MTSK-D3 (both 1.58), do equally well, marking them as the best fuzzy methods and matching SMOreg's performance.

Our findings demonstrate that TSK functions set to degree 3 (D3) generally provide more precise predictions across the datasets compared to the degree 2 (D2) variant. However, there are cases such as the Wankara and the Diabetes datasets where both perform in a similar range and D2 slightly outperforming D3, suggesting that a higher complexity in rules may not always lead to better results and could start over-fitting after a point. The D3 configuration emerged as the top-performing fuzzy model for the Treasury, Mortgage, Wankara and Concrete datasets, and it was also the best overall model for the Concrete datasets. On the other
hand, the D2 variant successfully matched the state-of-the-art fuzzy models in the Treasury, Wankara and Mortgage datasets. In the ELE-2 dataset, the D3 variant outperforms the opaque model approaches.

In the Diabetes dataset, the HIT2-MTSK method performed competitively by being close to the results obtained by using the best fuzzy approach (0.68 vs 0.69). Adopting alternative optimization strategies or refining ACO could enhance the performance of our HIT2-MTSK approach further. It is worth noting that the Diabetes dataset was inherently the most limited in terms of number of input features and it is possible that HIT2-MTSK performs much better in more complex datasets.

However, accuracy is only one part of the story when it comes to fuzzy modelling. The other part - interpretability is just as important to the end system users. A characteristic rule generated by the HIT2-MTSK approach is similar in structure to those in a Mamdani Fuzzy Rule-Based System (FRBS). For instance, as shown in Eq.(\ref{cement_eq}), a sample rule can be read as:

\emph{If Cement is High AND BlastFurnaceSlag is High THEN Compressive
Strength is High.}

Even though there is an equation associated with this rule, this does not take away the explainability as we have ensured that the equation always points in the boundary of the chosen fuzzy set. This rule very closely resembles a traditional Mamdani rule -- in fact the rule linguistic representation is indistinguishable from a traditional Mamdani FRBS rule. However, as discussed earlier, the rule differs in how the output is defined. Instead of pointing to the centroid of the fuzzy set \emph{High}, the output in the HIT2-MTSK system is derived from an equation involving the antecedent features, allowing it to represent any part of the consequent fuzzy set. This distinction ensures that the rules align more naturally with the linguistic meaning of terms like \emph{High}, making the rules not only more interpretable but also more precise than those in conventional Mamdani FRBSs.

\section{Case Study: Predicting California Housing Prices}
The effectiveness of the proposed method can be demonstrated through its application to the California Housing dataset, a widely used benchmark in machine learning and predictive modelling research. This dataset, originally derived from the 1990 U.S. Census, contains information about housing prices and their relationship to demographic and geographic features across various districts in California. The proposed HIT2-MTSK method will be compared against benchmark results including results from a Mamdani IT2 regression model - demonstrating the value that HIT2-MTSK adds over the Mamdani model. 
The task consists of predicting the median house value in each district based on explanatory variables such as median income, average number of rooms, average household size, population, and geographical coordinates. This prediction problem is not only of academic interest but also of practical relevance, as accurate housing price estimation supports decision-making in areas such as urban planning, real estate investment, and policy design.
The California Housing dataset is particularly suitable for evaluating the proposed method for two main reasons. First, it is a real-world dataset with inherent complexity, including non-linear relationships, heterogeneous feature distributions, and geographic variability. Second, the dataset has been extensively studied in the literature, which allows for a meaningful comparison with baseline models and previously reported results.
In the following subsections, we describe the dataset characteristics, experimental setup, results, and discussion of findings, including a comparison with the Mamdani approach applied to the same dataset, ultimately demonstrating the practical applicability of the proposed method in a real-world prediction task.
\subsection{Dataset Description}
The California Housing dataset originates from the 1990 U.S. Census and has become a standard benchmark for regression tasks in machine learning. It contains \num{20640} observations, each representing a California district, and is described by \num{8} predictive features along with one target variable. The features include both demographic and geographic attributes, which are summarized as follows:
\begin{itemize}
    \item \textbf{MedInc}: median income in the district (in tens of thousands of US dollars).
    \item \textbf{HouseAge}: median age of houses in the district.
    \item \textbf{AveRooms}: average number of rooms per household.
    \item \textbf{AveBedrms}: average number of bedrooms per household.
    \item \textbf{Population}: total population of the district.
    \item \textbf{AveOccup}: average household size (occupants per household).
    \item \textbf{Latitude}: geographical latitude of the district.
    \item \textbf{Longitude}: geographical longitude of the district.
\end{itemize}
The target variable is the \textbf{median house value} in each district, expressed in units of \num{100000} US dollars. 
Observations were randomly partitioned into training and testing subsets, with \num{80}\% of the data used for training and \num{20}\% reserved for evaluation. This partitioning ensures that the model is evaluated on unseen samples, providing a reliable measure of generalization performance.
\subsection{Results and Discussion}
Table~\ref{tab:results} presents the predictive performance of the HIT2-MTSK method in comparison with the baseline models on the California Housing dataset. Results are reported for  Root Mean Squared Error(RMSE). 
\begin{table}[h!]
\centering
\caption{Comparison of predictive performance on the California Housing dataset.}
\label{tab:results}
\begin{tabular}{lccc}
\hline
\textbf{Model} & \textbf{RMSE}  \\
\hline 
Linear Regression (LR)        & 0.728   \\
CART (RF)                      & 0.720   \\
NAM                            & 0.562   \\
EBM                            & 0.557   \\
XGBoost                         & 0.532 \\
DNN                             & 0.492 \\
Mamdani FRBS                    & 0.751 \\
HIT2-MTSK                    & 0.695 \\
\hline
\end{tabular}
\end{table}
The results demonstrate that the HIT2-MTSK method outperforms the other explainable baseline models on RMSE, including Mamdani FRBS. Compared to traditional linear regression, the reduction in error highlights the ability of the method to capture non-linear relationships in the data. Opaque approaches deliver a lower RMSE, however lack explainability at the level of training data as well as individual inferences.
\subsection{Evaluation Metrics for Explainability}
Drawing from the comprehensive evaluation framework proposed by Vilone and Longo~\citep{vilone_quantitative_2021}, we selected metrics applicable to direct fuzzy regression systems. Whilst their framework targets post-hoc wrapper models that explain black-box systems, several of their metrics - particularly those assessing rule complexity, coverage, and robustness - translate directly to evaluating the inherent interpretability of our fuzzy ruleset.
The HIT2-MTSK methodology was examined using five key metrics:
\textbf{Classes Covered} measures whether rules exist for different output ranges. Based on the prediction range in the test data (0.78 to 4.93), the model covers 96\% of the actual test data range (0.15 to 5.00), with slight compression at the extremes. The model achieved 100\% coverage in terms of the output fuzzy sets, with rules defined for Low, Medium, and High house value ranges.
\textbf{Active Rules per Prediction} counts rules with firing strength exceeding various thresholds. At a threshold of 0.15, an average of 8.38 rules fire per prediction. This drops to 6.33 at 0.25 and 3.83 at 0.5. The decrease at higher thresholds shows that whilst multiple rules contribute to each prediction, only a few dominate with high firing strength.
\textbf{Rule Base Characteristics} comprise two measures. The rule base contains 75 rules total, each with an average of 2.67 antecedents. These values indicate a compact, interpretable model where individual rules remain simple enough for human comprehension.
\textbf{Dataset Coverage} confirms that 100\% of test instances activate at least one rule. This complete coverage ensures the model can handle all input combinations without gaps.
\textbf{Robustness to Noise} was assessed by adding Gaussian noise at three levels to input features. The noise was added proportionally based on the standard deviations of each feature in the training dataset. The mean absolute change in predictions, expressed as a percentage of the mean house value, was 1.18\% at 1\% noise, 5.84\% at 5\% noise, and 12.24\% at 10\% noise levels. The near-linear relationship between noise level and prediction change demonstrates stable behaviour without sudden discontinuities.
These metrics collectively demonstrate that HIT2-MTSK maintains the interpretability expected of fuzzy systems whilst achieving comprehensive coverage and robust predictions. The gradual decrease in active rules at higher firing thresholds shows that whilst multiple rules contribute to each prediction, a smaller subset dominates the decision-making process.
\subsection{Comparison with Mamdani FRBS}
\begin{figure}[h!]
    \centering
    \includegraphics[width=0.65\linewidth]{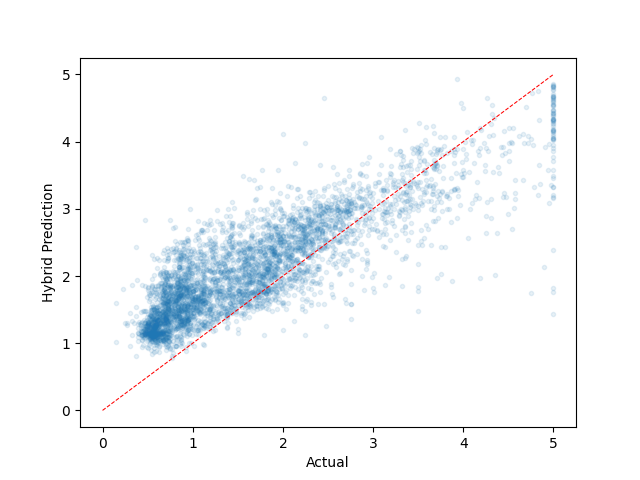}
    \caption{Scatter plot of actual versus predicted median house values using HIT2-MTSK}
    \label{fig:scatter_hybrid}
\end{figure}

\begin{figure}[h!]
    \centering
    \includegraphics[width=0.65\linewidth]{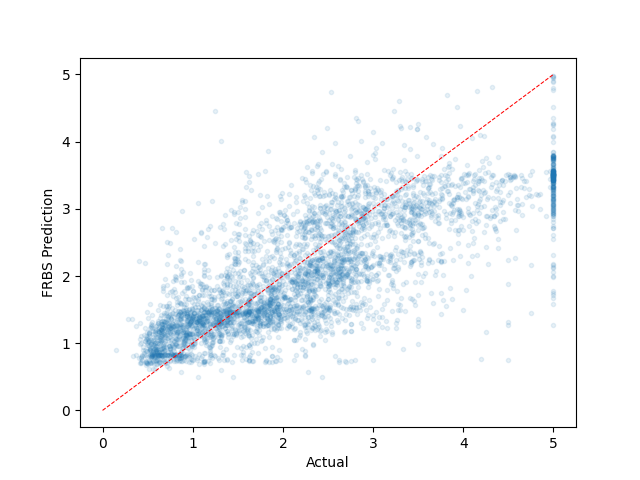}
    \caption{Scatter plot of actual versus predicted median house values using a Mamdani FRBS}
    \label{fig:scatter_mamdani}
\end{figure}

A primary goal of this research was to surpass Mamdani FRBS performance while preserving its level of explainability. The HIT2-MTSK models achieved their results using a compact rule-base containing 75 rules. Although Mamdani FRBS models were trained with the same number of rules, they struggled to achieve comparable predictive accuracy. To ensure fair comparison, both models used identical configurations: the same number of input and output fuzzy-sets and the same maximum rule length.
Figure~\ref{fig:scatter_hybrid} displays the relationship between predicted and actual house values for the testing dataset using the HIT2-MTSK approach. Figure~\ref{fig:scatter_mamdani} shows the corresponding scatter plot for the same data using the FRBS model.
These figures clearly demonstrate that the hybrid approach produces a much tighter clustering around the ideal prediction line. The Mamdani FRBS scatter plot reveals distinct horizontal banding patterns at positions corresponding to the output fuzzy set centroids, visible as horizontal structures in Figure~\ref{fig:scatter_mamdani}. This banding occurs in Mamdani systems when all (or most) rules for a given prediction activate the same output fuzzy-set, causing predictions to cluster within a narrow range. The hybrid approach eliminates this effect by generating predictions across the entire support region of the output fuzzy-sets, resulting in a more uniform distribution of predicted values.

\begin{figure}[h!]
    \centering
    \includegraphics[width=0.65\linewidth]{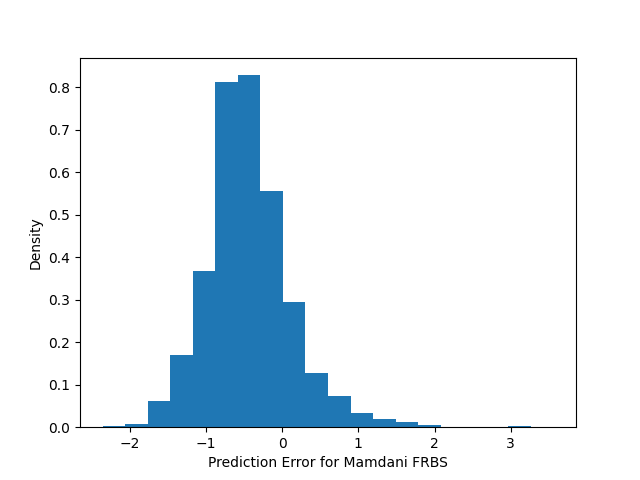}
    \caption{Residual distribution for HIT2-MTSK}
    \label{fig:residual_hybrid}
\end{figure}

\begin{figure}[h!]
    \centering
    \includegraphics[width=0.65\linewidth]{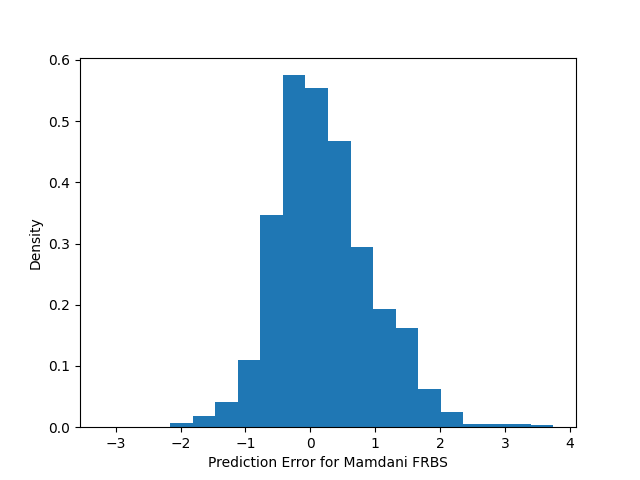}
    \caption{Residual distribution for Mamdani FRBS}
    \label{fig:residual_mamdani}
\end{figure}

The residual distribution for the HIT2-MTSK approach (Figure~\ref{fig:residual_hybrid}) shows errors distributed approximately symmetrically around zero, indicating no significant systematic bias. The Mamdani approach's residual distribution (Figure~\ref{fig:residual_mamdani}) similarly shows no bias, but exhibits a wider error spread, confirming the superior performance of the hybrid HIT2-MTSK approach.

The experimental results confirm the proposed method's effectiveness in predicting housing prices and demonstrate clear improvements over the traditional Mamdani approach. Its superior performance compared to established baselines confirms both its robustness and practical applicability for real-world regression tasks.
\section{Conclusions and Future Work}
\label{conclusions-and-future-work}
The primary innovation of this research lies in the modifications to the rule structure, which enhanced both accuracy and interpretability. However, some limitations are evident as the performance can be improved further on some datasets. Future research could focus on optimization enhancements through other approaches, such as meta-heuristic techniques, and domain-specific adaptations like tailored pre-processing and feature selection. Expanding the evaluation to include additional datasets and alternative metrics will further validate the approach and support its refinement.
In summary, the HIT2-MTSK approach demonstrates significant improvements over the traditional Mamdani FRBS. It performs well in 5 out of 6 of the benchmark datasets achieving either state-of-the-art results or outperforming the majority of benchmark algorithms. Additionally, the hybrid fuzzy methodology strikes an effective balance between accuracy and interpretability, which could potentially fill a crucial gap in current fuzzy modelling techniques.

\bibliography{zotero_bib_4}

\end{document}